\documentclass{article}

\usepackage{PRIMEarxiv}

\usepackage{hyperref}       % Hyperlinks
\usepackage{url}            % URL formatting
\usepackage{booktabs}       % Three-line tables
\usepackage{amsfonts}       % Mathematical symbols
\usepackage{amssymb}        % AMS symbols (checkmark)
\usepackage{amsmath}        % Mathematical formulas
\usepackage{pifont}         % Dingbat symbols (ding{55})
\usepackage{microtype}      % Font microtypography
\usepackage{fancyhdr}       % Headers and footers
\usepackage{graphicx}       % Graphics support
\graphicspath{{figures/}}   % Graphics path
\usepackage{xcolor}         % Color support
\usepackage{listings}       % Code rendering
\usepackage{enumitem}       % List formatting
\usepackage{caption}        % Figure/table captions
\usepackage{subcaption}     % Subfigure support

\definecolor{blueaccent}{HTML}{1a73e8}
\definecolor{redalert}{HTML}{d93025}
\definecolor{greenok}{HTML}{188038}
\definecolor{codegray}{rgb}{0.5,0.5,0.5}
\definecolor{codepurple}{rgb}{0.58,0,0.82}

\hypersetup{
    colorlinks=true,
    linkcolor=blueaccent,
    citecolor=blueaccent,
    urlcolor=blueaccent,
}

\title{Constraint Tax in Open-Weight LLMs: An Empirical Study of Tool Calling Suppression Under Structured Output Constraints}

\author{
Fangzheng Li$^{1,2,\dagger}$ \\
$^{1}$Focus AI Center, Focus Technology Co., Ltd. \\
$^{2}$Nanjing University of Science and Technology \\
\texttt{lifangzheng@focuschina.com} \\
\texttt{Fangzheng\_Li@njust.edu.cn}
\and
Aimin Zhang$^{1,\dagger}$\thanks{Corresponding author: zhangaimin@focuschina.com} \\
$^{1}$Focus AI Center, Focus Technology Co., Ltd. \\
\texttt{zhangaimin@focuschina.com}
\and
Chen Lv$^{1}$ \\
$^{1}$Focus AI Center, Focus Technology Co., Ltd. \\
\texttt{lvchen1018@focuschina.com}
}

\begin{document}
\maketitle

\begin{abstract}
Tool Calling and Structured Output are two core capabilities of modern Agent systems, yet their interaction under joint deployment conditions remains insufficiently understood. This paper reports a reproducible phenomenon observed in a production Agent system: when Tool Calling and JSON Schema constraints are simultaneously enabled, multiple open-weight models cease invoking tools despite maintaining high schema compliance. We refer to this behavior as \textbf{Tool Suppression}.

Through controlled experiments across multiple model families and deployment settings, we consistently reproduce Tool Suppression under joint constraints, while tool execution and schema compliance remain functional when evaluated independently. Further analysis reveals that JSON Schema constraints are compiled into grammar-based token masks, causing tool-call tokens to become unreachable during decoding. This provides an implementation-level explanation for the observed behavior.

To interpret the phenomenon, we formulate the \textbf{Constraint Priority Inversion (CPI)} hypothesis, which suggests that schema satisfaction may dominate action-selection behavior under multiple simultaneous constraints. We present CPI as a behavioral hypothesis consistent with the observed evidence rather than a verified internal mechanism.

To mitigate the problem, we propose \textbf{Transparent Two-Pass Execution}, an inference-time strategy that decouples tool execution from schema-constrained response generation. Experimental results show that this approach restores tool invocation while preserving structured output guarantees without requiring model retraining.

These findings suggest that evaluating tool use and structured output separately may overlook important reliability issues in production Agent systems. Code, data, and docs will be released at https://github.com/Fzsama/Constrain-Tax-26-06.git.
\end{abstract}

% keywords

% ===== 1. Introduction =====
\section{Introduction}

\subsection{Tool-Augmented LLM Agents}

As Large Language Models (LLMs) evolve from text-only interfaces to action-taking agents, tool augmentation has become a standard paradigm for enabling models to interact with external environments. With the development of standardized protocols such as MCP and OpenAI-compatible tool-calling APIs, tool execution has been widely adopted in production Agent systems.

Meanwhile, Structured Output has become another critical capability in production Agent systems. Rather than returning unconstrained natural language, deployed Agents are often required to generate responses that satisfy predefined JSON schemas for downstream parsing, workflow automation, API integration, and multi-Agent coordination.

As a result, Tool Calling and Structured Output are increasingly activated simultaneously within the same execution pipeline. A typical production workflow may require an Agent to first acquire external information through tools and then organize the retrieved information into a schema-compliant response.

Although both capabilities have been extensively studied individually, relatively little attention has been paid to their interaction under joint deployment conditions. Existing evaluations typically assess tool use, structured generation, and task completion separately, implicitly assuming that capabilities that function correctly in isolation will continue to function correctly when combined.

Whether this assumption holds in production Agent environments remains insufficiently understood and motivates the investigation presented in this paper.

\subsection{Constraint Tax and an Unexplored Agent Failure Mode}

Recent research has begun to focus on the unintended effects of structured-output constraints on model behavior. This line of work has observed that requiring models to generate outputs in specific formats can degrade answer quality, reduce factual accuracy, and increase token usage. This phenomenon has been described as a ``Constraint Tax'' on model performance in text-generation tasks.

This finding indicates that structured-output constraints are not computationally neutral; they impose measurable costs on model behavior beyond format compliance.

Agent systems introduce a fundamentally different execution setting. Unlike traditional text-generation tasks, Agent workflows require models to make decisions about whether external actions should be executed, when they should be executed, and how externally acquired information should be incorporated into the final response.

Consequently, when Tool Calling and Structured Output constraints coexist, structured generation constraints may influence not only the form of the final response but also the execution process itself. One possible mechanism is that decoding-level constraint enforcement—such as grammar-based token masking—may render tool-call tokens unreachable during generation, thereby preventing tool execution at the output layer.

This observation motivates an open research question:

\begin{quote}
When Tool Calling and Structured Output constraints are simultaneously enabled, do they interact in ways that affect Agent execution behavior? If so, at what level does this interaction occur—model preference, decoding constraint, or both?
\end{quote}

To our knowledge, this question has received limited empirical investigation. Existing Constraint Tax studies primarily focus on answer quality degradation under structured-output requirements, whereas the potential impact on Agent action execution and the underlying implementation-level mechanisms remain largely unexplored.

This paper investigates that question through controlled experiments, inference-stack tracing, and production-system observations.

\subsection{Observation from a Production Agent System}

The investigation was initially motivated by an unexpected observation during the deployment of an Agent system in a production environment. The system was configured to use an open-weight model with both Tool Calling and Structured Output constraints enabled, a configuration that is common in production settings.

Under this joint-constraint configuration, the Agent repeatedly failed to invoke external tools even when tasks explicitly required external information acquisition. Tool call events were not generated, and the final response was produced without incorporating any external data.

In contrast, when the Structured Output constraint was disabled while keeping all other conditions unchanged, the same model successfully invoked tools and completed the tasks.

This behavior was unexpected from a system-design perspective because the only modification involved the presence of a schema constraint, while model weights, prompts, tool definitions, and task requirements remained identical.

The key observation was that the model appeared to satisfy the schema requirement while silently bypassing the tool execution step entirely. This pattern was initially treated as an implementation anomaly but persisted across repeated trials.

The observation described above motivates the central question investigated in this paper. Both Tool Calling and Structured Output generation remain functional when evaluated independently. However, under joint deployment conditions, tool execution behavior may disappear despite the continued presence of tool definitions, unchanged task requirements, and successful schema generation.

Understanding the origin of this behavioral inconsistency forms the primary objective of the study.

\subsection{Contributions}

This paper makes the following contributions:

\begin{itemize}

\item \textbf{Identification and characterization of Tool Suppression.}

We report a reproducible failure pattern observed in a production Agent system in which multiple evaluated open-weight models cease invoking tools when Tool Calling and Structured Output constraints are simultaneously enabled. We define this behavior as \emph{Tool Suppression} and characterize its observable properties through controlled experiments.

\item \textbf{Root cause localization to grammar-based constrained decoding.}

Through systematic tracing of the inference stack, we identify that JSON Schema constraints are compiled into grammar-based token masks that render tool-call tokens unreachable during decoding. This provides a concrete implementation-level explanation for the observed suppression phenomenon.

\item \textbf{Formulation of the Constraint Priority Inversion (CPI) hypothesis.}

Based on behavioral evidence collected during diagnostic experiments, we introduce CPI as a possible interpretation of the observed suppression pattern. CPI is presented as a behavioral hypothesis consistent with the observed evidence rather than a verified internal mechanism.

\item \textbf{Proposal and validation of Transparent Two-Pass Execution.}

We propose an inference-time mitigation strategy that separates tool execution from schema-constrained response generation. Experimental results show that the approach restores tool invocation behavior while preserving structured output guarantees.

\item \textbf{Development of a Tool Suppression behavioral taxonomy.}

We further summarize recurring suppression patterns into a behavioral taxonomy (TS-A to TS-E), providing a descriptive framework for analyzing suppression behavior across different models and deployment settings.

\end{itemize}

% ===== 2. Background =====
\section{Background}

\subsection{Tool Calling in LLM Agents}

Tool Calling refers to the ability of language models to interact with the external environment by generating structured function call requests. Unlike traditional language models that rely solely on parametric knowledge, tool-augmented Agents can access external resources such as search engines, databases, code interpreters, and enterprise internal systems, thereby completing tasks that require real-time information acquisition, environmental perception, or external execution~\cite{yao2022react,schick2023toolformer,qin2023tool}. Therefore, tool calling has become one of the important features distinguishing modern Agent systems from pure dialogue models.

Existing Agent frameworks commonly treat tool calling as an execution process composed of multiple stages, including Task Understanding, Tool Planning, Tool Selection, and Tool Execution. In this process, models not only need to determine whether external information is required but also need to generate correct tool call actions and utilize returned results for subsequent reasoning. In recent years, extensive research has been conducted on tool learning capabilities, including evaluation metrics such as Tool Selection Accuracy, Task Completion Rate, and Tool Use Success Rate~\cite{qin2023tool,lin2026scaling}.

Existing research has identified certain limitations of open-weight models in tool use. Shen et al.~\cite{shen2024small} pointed out that small-scale language models exhibit significant disadvantages in tool learning tasks and proposed the ``Weak Tool Learners'' phenomenon. Wang et al.~\cite{wang2024learning} further demonstrated that incorporating failure cases during Agent fine-tuning can effectively improve tool use behavior. These works indicate that tool calling capability itself has become an important research direction in Agent research.

However, existing research primarily focuses on whether tool calling capability exists and whether tool calling is correctly executed, with less attention to the interaction effects between tool calling capability and other system constraints. Particularly in scenarios where tool calling and structured output constraints coexist, whether models can maintain normal tool execution behavior remains systematically understudied. The Tool Suppression phenomenon studied in this paper occurs precisely in this joint constraint scenario.

\subsection{Structured Output Generation}

Structured Output Generation refers to the technical paradigm of constraining language model output format through predefined Schemas, enabling generated results to be stably parsed and consumed by programs. Unlike traditional natural language responses, structured output requires models to organize content according to specific data structures, typically manifested as JSON objects, function parameters, or text conforming to formal grammar constraints. As LLMs are increasingly integrated into production systems, structured output has become a fundamental capability for Agents, workflow orchestration systems, and multi-Agent collaboration frameworks.

Current mainstream implementation approaches mainly include two categories. The first is API-layer constraints, such as the \texttt{response\_format} mechanism in OpenAI-compatible interfaces, which guides models to generate output structures meeting requirements through Schema descriptions. The second is Constrained Decoding at the decoding layer, which restricts the search space during token generation through grammar constraints, state machines, or finite automata, such as \texttt{guided\_json} in SGLang and Grammar-based Decoding mechanisms in vLLM. Regardless of the implementation approach, the core objective is to improve the determinability and parsability of output format.

Existing research generally treats structured output as an engineering reliability mechanism. Liu et al.~\cite{liu2024structured} found through user surveys that structured output requirements have become a prevalent demand in industrial scenarios; Deng et al.~\cite{deng2025decoupling} further demonstrated that there exists a significant coupling relationship between output format and task solving process, and directly imposing format constraints on the generation process may affect the model's ability to complete the task itself.

This finding resonates with Constraint Tax research. Growing evidence suggests that structured output constraints do not only act on the final output stage but may alter the model's resource allocation and decision behavior during reasoning. When models simultaneously need to satisfy content generation objectives and format constraint objectives, competition may arise between the two types of objectives. However, existing research primarily focuses on the impact of structured output on answer quality, reasoning capability, and format compliance rate, with insufficient empirical research on whether it further intervenes in action selection behavior during Agent execution.

The Tool Suppression phenomenon focused on in this paper occurs precisely in this context: structured output constraints no longer merely affect ``how to answer'' but may further affect whether the Agent ``executes actions.''

\subsection{Constraint Tax}

In recent years, with the widespread application of structured output in industrial scenarios, researchers have begun to focus on the impact of format constraints on model behavior itself. Ray~\cite{ray2026constraint} first systematically proposed the \textbf{Constraint Tax} concept to describe the performance cost that language models pay when satisfying structured output requirements. Experimental results demonstrate that when strict Schema constraints are imposed on the generation process, models can achieve higher format compliance rates, but their answer accuracy may significantly decrease.

This phenomenon indicates that structured output constraints are not free engineering capabilities. Traditional views typically assume that Schema constraints only affect output form without affecting the model's ability to solve tasks themselves. However, Constraint Tax research shows that format constraints actually participate in decision competition during the model generation process: models not only need to complete task solving but also simultaneously satisfy format compliance objectives.

From a more general perspective, the Constraint Tax can be understood as a \textbf{Goal Competition} phenomenon. When models face multiple simultaneous optimization objectives, their limited reasoning resources and generation capabilities need to be allocated across different objectives~\cite{agyemang2026ogc}. For example, models may simultaneously need to:

\begin{itemize}
    \item Generate correct content (Content Correctness);
    \item Satisfy format requirements (Format Compliance);
    \item Invoke external tools (Tool Execution);
    \item Control generation length (Length Constraints).
\end{itemize}

Ideally, these objectives should be simultaneously satisfied; however, in actual models, different objectives may form competitive relationships. When one objective receives excessively high priority, other objectives may be partially or even completely sacrificed. The Constraint Tax is precisely a typical manifestation of this goal competition in structured output scenarios.

Existing research primarily examines the impact of the Constraint Tax on answer accuracy and reasoning capability, so its manifestation is typically described as ``format-correct but answer-incorrect.'' However, for Agent systems, content generation is only the final stage in the execution pipeline. In tool-augmented Agents, models also need to complete action decision processes such as tool planning, tool selection, and tool execution. Therefore, a natural but insufficiently studied question is: do structured output constraints further affect the Agent's action selection behavior, rather than merely affecting final answer quality?

This paper's research demonstrates that such influence indeed exists. In Agent scenarios, the Constraint Tax no longer only manifests as content quality degradation but may further evolve into systematic absence of tool execution behavior. We refer to this phenomenon as \textbf{Tool Suppression} and propose Constraint Priority Inversion (CPI) as a possible explanation mechanism in subsequent sections.

\subsection{Grammar-Constrained Decoding}

Grammar-constrained decoding is a technique used in modern inference frameworks to enforce structured output formats at the token-generation level. Rather than relying solely on model prompting or post-hoc validation, this approach actively constrains the set of tokens that the model may generate at each decoding step.

Frameworks such as SGLang and vLLM integrate grammar-based constraint engines including xgrammar and Outlines. These engines compile user-provided JSON Schemas into finite-state machines (FSMs) that define the set of valid token sequences. At each decoding step, the FSM determines which tokens are permissible given the current generation state, and the inference framework applies a vocabulary mask to set impermissible token logits to \(-\infty\), effectively making them impossible to sample.

This mechanism ensures format compliance with high reliability, which is particularly valuable in production environments where downstream systems require deterministic parsing. However, the application of grammar constraints is not limited to the final output stage; it operates at the token level throughout generation.

The effect of grammar-constrained decoding on token availability has been characterized in prior work on constrained generation. For example, Garbacea and Mei~\cite{garbacea2022constrained} discuss the structural challenges introduced by vocabulary restrictions in constrained language generation. Suresh et al.~\cite{suresh2025dingo} further examine the interaction between constrained inference and model behavior. These works suggest that grammar constraints can fundamentally alter token availability during generation, which may have implications beyond simple formatting compliance.

In the context of Agent systems that use XML-style tool-call formats, this token-level constraint mechanism introduces an important consideration: when a JSON Schema grammar is active, any token sequence that deviates from the grammar—including tool-call tags—becomes unreachable. This creates a potential tension between structured output requirements and tool execution capability at the decoding layer.

\subsection{Agent Execution Pipelines}

Tool-augmented Agents are typically modeled as execution pipelines composed of multiple consecutive decision stages. From a functional perspective, Agents not only need to generate natural language responses but also need to complete a series of intermediate steps such as task analysis, tool decision-making, and external environment interaction. Therefore, the Agent execution process is essentially a Multi-stage Decision Making process rather than a pure text generation task.

Although existing Agent frameworks differ in specific implementations, their execution logic can generally be abstracted as the following pipeline:

\begin{center}
\texttt{User Query}
$\rightarrow$
\texttt{Task Understanding}
$\rightarrow$
\texttt{Tool Planning}
$\rightarrow$
\texttt{Tool Execution}
$\rightarrow$
\texttt{Response Generation}
\end{center}

Where:

\begin{itemize}
    \item \textbf{Task Understanding}: The model identifies user intent, task objectives, and required information;
    
    \item \textbf{Tool Planning}: The model determines whether external tools are needed and formulates tool usage plans;
    
    \item \textbf{Tool Execution}: The model initiates tool calls and receives external return results;
    
    \item \textbf{Response Generation}: The model generates final responses based on tool results.
\end{itemize}

Existing Agent evaluation benchmarks mostly focus on tool planning and tool execution capabilities. For example, works such as ToolBench, API-Bank, and BMTools primarily evaluate whether models can correctly select tools and whether they can complete tasks~\cite{qin2023tool,wang2023survey}. Meanwhile, structured output-related research mainly focuses on format compliance during the response generation stage.

However, this evaluation approach implicitly assumes that different stages in the Agent execution pipeline are independent, i.e., constraints acting on the response generation stage do not affect the decision process of preceding stages. In existing literature, this assumption is rarely explicitly validated.

The observations in this paper indicate that this assumption may not hold. When structured output constraints are imposed on the response generation stage, their influence may propagate backward to the tool execution stage, thereby changing the Agent's action selection results. In other words, a constraint that originally belongs to the output level may affect execution-level behavior.

Based on this perspective, this paper positions Tool Suppression as a Cross-Stage Interference phenomenon in the Agent execution pipeline: task understanding and tool planning processes still complete normally, but the tool execution stage is systematically skipped, while the response generation stage continues to produce format-compliant output. Subsequent sections will further demonstrate that this behavioral pattern is fundamentally different from traditional tool capability deficiency.

% ===== 3. Problem Definition =====
\section{Problem Definition}

\subsection{Constraint Tax in Agent Contexts}

Ray~\cite{ray2026constraint} defines the Constraint Tax as the performance cost introduced when language models are required to satisfy additional structured-output constraints under fixed model and task conditions. Existing studies primarily observe this cost through reduced answer quality, degraded reasoning performance, or increased proportions of format-compliant but content-incorrect outputs.

From a broader perspective, Constraint Tax can be viewed as a behavioral consequence of operating under multiple simultaneous objectives. In addition to solving the primary task, models may also be required to satisfy format constraints, safety requirements, workflow specifications, or other deployment-related conditions.

For Agent systems, these objectives coexist with action-oriented behaviors such as tool planning and tool execution. Consequently, structured-output constraints may influence not only response generation but potentially other stages of Agent execution as well.

The present study investigates this possibility in the context of Tool Calling and Structured Output constraints operating simultaneously within production Agent environments. As we will show, the interaction may originate not only from model-level behavioral tendencies but also from decoding-level constraint enforcement mechanisms.

\subsection{Tool Suppression}

We define the core phenomenon studied in this paper as \textbf{Tool Suppression}.

\begin{quote}
\textbf{Tool Suppression} refers to the phenomenon where, in tasks with genuine tool requirements, models possess tool calling capability and can identify tool usage needs, but systematically abandon tool execution behavior under joint constraint conditions.
\end{quote}

This definition includes three necessary conditions:

\begin{enumerate}

\item \textbf{Tool Requirement}. The task itself requires accessing external information beyond the model's parameters, or requires solving through external tools.

\item \textbf{Tool Capability}. The model can normally complete tool calling under conditions without joint constraints.

\item \textbf{Execution Omission}. The model ultimately does not initiate tool calls but directly generates responses.

\end{enumerate}

It is worth noting that Tool Suppression is fundamentally different from traditional Tool Incapability. In Tool Incapability scenarios, models cannot correctly identify tool requirements or cannot generate valid tool calls; whereas in Tool Suppression scenarios, models possess the corresponding capability, but this capability is not activated under specific constraint conditions.

From the Agent execution pipeline perspective, Tool Suppression manifests as a Selective Failure:

\[
\text{Task Understanding}
\;\checkmark
\]

\[
\text{Tool Planning}
\;\checkmark
\]

\[
\text{Tool Execution}
\;\times
\]

\[
\text{Response Generation}
\;\checkmark
\]

That is, models can understand tasks, identify information gaps, and generate final responses, but exhibit systematic skipping during the tool execution stage.

This phenomenon has a similar structure to the ``format-correct but answer-incorrect'' in traditional Constraint Tax: both manifest as constraint conditions causing degradation of certain core capabilities. However, in Tool Suppression, the degraded object is no longer answer quality but the Agent's action execution capability. Therefore, we regard it as a Behavior-Level Manifestation of the Constraint Tax in Agent systems.

Subsequent sections will further use Tool Invocation Rate (TIR) and Suppression Rate (SR) to quantitatively characterize this phenomenon.

\subsection{Tool Invocation Rate}

To quantitatively characterize tool calling behavior, we define the \textbf{Tool Invocation Rate (TIR)} as follows:

\begin{equation}
TIR =
\frac{N_{\text{tool}}}
     {N_{\text{total}}}
\end{equation}

Where:

\begin{itemize}
    \item $N_{\text{tool}}$ represents the number of requests that successfully initiated at least one tool call;
    \item $N_{\text{total}}$ represents the total number of requests.
\end{itemize}

Therefore, $TIR \in [0,1]$, with higher values indicating that the model is more inclined to execute tool calling behavior.

It should be emphasized that TIR measures \textbf{Tool Execution Behavior}, not tool calling quality. As long as the model successfully triggers at least one valid tool call, it counts as one tool execution event; whether the tool returns correct results and the quality of the final answer do not affect TIR statistics.

This design is consistent with this paper's research objectives. Since Tool Suppression focuses on ``whether to execute tools'' rather than ``whether to correctly use tools,'' TIR can directly reflect whether Agent execution behavior is suppressed.

Ideally, for a task set with genuine tool requirements, we should have:

\begin{equation}
TIR_{\text{baseline}}
\approx
1
\end{equation}

That is, the model can stably execute tool calls.

If after introducing joint constraints we observe:

\begin{equation}
TIR_{\text{constrained}}
\ll
TIR_{\text{baseline}}
\end{equation}

This indicates that tool execution behavior is significantly affected. In extreme cases, when

\begin{equation}
TIR_{\text{constrained}}
=
0
\end{equation}

This means the model completely stops tool calling behavior, manifesting as Complete Tool Suppression.

\subsection{Suppression Rate}

Although TIR can reflect the model's tool execution behavior, its absolute value is difficult to directly characterize the additional impact caused by joint constraints. For example, different models may inherently have different tool calling tendencies, so comparing only TIR under constraint conditions cannot accurately measure the severity of Tool Suppression.

To this end, we further define the \textbf{Suppression Rate (SR)}:

\begin{equation}
SR
=
1
-
\frac{TIR_{\text{constrained}}}
     {TIR_{\text{baseline}}}
\end{equation}

Where:

\begin{itemize}
    \item $TIR_{\text{baseline}}$ represents the tool invocation rate without joint constraints;
    \item $TIR_{\text{constrained}}$ represents the tool invocation rate with joint constraints.
\end{itemize}

When $TIR_{\text{baseline}} > 0$,

\[
SR \in [0,1]
\]

Higher values indicate stronger suppression of tool calling behavior.

According to this definition:

\begin{itemize}

\item When

\[
TIR_{\text{constrained}}
=
TIR_{\text{baseline}}
\]

then

\[
SR = 0
\]

indicating that joint constraints have no observable impact on tool execution behavior;

\item When

\[
0
<
TIR_{\text{constrained}}
<
TIR_{\text{baseline}}
\]

then

\[
0
<
SR
<
1
\]

indicating Partial Tool Suppression;

\item When

\[
TIR_{\text{constrained}}
=
0
\]

then

\[
SR = 1
\]

indicating Complete Tool Suppression.

\end{itemize}

Compared to directly using TIR, SR can eliminate the influence of the model's inherent tool calling tendency, making it more suitable as a standardized indicator for cross-model comparison and constraint effect analysis.

In summary, Tool Suppression defines the core phenomenon studied in this paper, while TIR and SR provide quantifiable measurement methods from behavioral and effect levels respectively. In subsequent experiments, this paper uses TIR to characterize tool execution behavior and SR to measure the degree of suppression caused by joint constraints.

\subsection{Taxonomy of Tool Suppression Behaviors}

This taxonomy is constructed based on three dimensions:
\begin{itemize}
\item \textbf{Tool Awareness}: Whether the model identifies the need for external tools;
\item \textbf{Tool Execution}: Whether the model actually initiates tool calls;
\item \textbf{Content Generation Strategy}: How the model constructs responses when unable to obtain tool results.
\end{itemize}
Based on the above dimensions, we propose the following taxonomy.
\paragraph{TS-A: Empty Compliance}
The model's generated response fully complies with the required schema, but key fields contain null values, default values, or placeholders.
\begin{lstlisting}
{"recommendations": [], "key_findings": []}
\end{lstlisting}
This pattern indicates that the model minimizes content generation workload while prioritizing format compliance. This phenomenon was mainly observed in GPT-OSS-20B.
\paragraph{TS-B: Simulated Retrieval}
The model generates information that appears to originate from external tools without ever calling any tools.
\begin{lstlisting}
{"buyer_background":
"Simulated Web Search Results"}
\end{lstlisting}
Compared to TS-A, this pattern actively fabricates content, generating data that appears to be retrieval results. Since its output is superficially plausible, downstream systems have greater difficulty detecting it. This behavior was mainly observed in Qwen3.5-122B-A10B.
% \paragraph{TS-C: Intent Without Action}
% The model explicitly expresses the need to use tools but ultimately fails to execute the corresponding operations.
% \begin{lstlisting}
% {
% "need_search": true,
% "tool_requirements": [...]
% }
% \end{lstlisting}
% This category provides the most compelling diagnostic evidence because it demonstrates a dissociation between tool awareness and tool execution. In Experiment B, all five trials produced \texttt{need\_search=true}, but none resulted in actual tool calls.
\paragraph{TS-C: Intent Without Action}

The model explicitly expresses the need to use tools but ultimately fails to execute the corresponding operations.

\begin{verbatim}
{
    "need_search": true,
    "tool_requirements": [...]
}
\end{verbatim}

This category is particularly informative because it reveals a behavioral separation between explicit recognition of tool requirements and the absence of actual tool execution.

In the evaluated examples, models occasionally produced outputs indicating awareness of external information requirements while no corresponding tool calls were observed. This pattern motivates further investigation into how action-selection decisions are made under joint constraints—specifically, whether the failure originates from model-level preferences, decoding-level constraints, or their interaction.

\paragraph{TS-D: Tool-Free Hallucination}
The model neither calls tools nor acknowledges insufficient information, but directly generates content lacking substantiation.
This pattern represents typical tool-free hallucination and is the highest-risk failure mode in deployment environments. Since the generated content is usually fluent and seemingly plausible, it may mislead users or downstream systems.
\paragraph{TS-E: Frozen Required Tool}
Even when explicitly forced to use tools (e.g., through \texttt{tool\_choice="required"} settings), the model still fails to call tools.
This behavior indicates that Tool Suppression cannot always be explained by incorrect assessment of tool requirements. Instead, it may reflect deeper failures in the action selection mechanism.
Table~\ref{tab:taxonomy} summarizes these five categories of Tool Suppression behavior.
\begin{table}[ht]
\centering
\caption{Taxonomy of Tool Suppression Behaviors}
\label{tab:taxonomy}
\begin{tabular}{lcccc}
\toprule
Category & Output Quality & Tool Awareness & Tool Execution & Deceptiveness \\
\midrule
TS-A Empty Compliance & Low & Absent & Blocked & Low \\
TS-B Simulated Retrieval & Medium & Absent & Blocked & Medium \\
TS-C Intent Without Action & Medium to High & \textbf{Present} & Blocked & Low \\
TS-D Tool-Free Hallucination & High & Absent & Blocked & \textbf{High} \\
TS-E Frozen Required Tool & Medium & Absent & Blocked (Forced) & Low \\
\bottomrule
\end{tabular}
\end{table}
The taxonomy presented above is intended as a descriptive framework rather than a causal explanation.

The five categories capture different observable manifestations of Tool Suppression across models and deployment settings. Some categories primarily reflect omission behavior (TS-A, TS-E), while others involve content-generation strategies that compensate for missing tool results (TS-B, TS-D). TS-C is distinctive because it exhibits explicit acknowledgement of tool requirements without corresponding execution behavior, motivating deeper investigation into the underlying constraint mechanisms. This taxonomy provides a common vocabulary for analyzing suppression behavior in subsequent experiments.

% ===== 4. Experimental Setup =====
\section{Experimental Setup}

\subsection{Research Questions}

This study investigates whether the coexistence of Tool Calling and Structured Output constraints introduces a systematic degradation in tool execution behavior for open-weight LLM agents.

Rather than evaluating general reasoning quality, we focus specifically on action execution under joint constraints.

The study addresses four research questions:

\begin{itemize}[nosep]
\item \textbf{RQ1 (Reproducibility)}: Can tool suppression be consistently reproduced across different open-weight model families under identical task settings?

\item \textbf{RQ2 (Constraint Interaction)}: Does the suppression behavior originate from the interaction between Tool Calling and Structured Output constraints rather than from either capability individually?

\item \textbf{RQ3 (Alternative Explanations)}: Can the observed phenomenon be explained by inference frameworks, deployment environments, model scale, quantization methods, schema design, or tool invocation settings?

\item \textbf{RQ4 (Mitigation)}: Can production-oriented engineering strategies restore tool execution behavior while preserving structured output guarantees?

\end{itemize}

The above research questions correspond to four levels: phenomenon verification, scale effect analysis, causal exclusion verification, and engineering mitigation exploration.

\subsection{Controlled Experimental Design}

To isolate the effect of joint constraints, we adopt a controlled three-condition design. The specific configurations and objectives for each condition are defined as follows:

\begin{itemize}
    \item \textbf{T1 (Tool-Only Baseline)}: Tools = ON, Response Format = OFF. This condition measures the model's baseline ability to invoke tools when no structured output constraints are present.
    \item \textbf{T2 (Joint Constraint Condition)}: Tools = ON, Response Format = ON. This condition represents the production setting where tool execution and JSON Schema compliance are simultaneously required.
    \item \textbf{T3 (Schema-Only Control)}: Tools = OFF, Response Format = ON. This condition verifies whether models can independently satisfy structured output constraints without tool execution requirements.
\end{itemize}

\begin{table}[ht]
\centering
\caption{T1/T2/T3 Experimental Conditions}
\label{tab:conditions}
\renewcommand\arraystretch{1.3} 
\begin{tabular}{lccl}
\toprule
Condition & Tools & Response Format & Purpose \\
\midrule
T1 (Baseline) & ON & OFF & Measure baseline tool calling capability \\
T2 (Joint Constraint) & ON & ON & Detect Tool Suppression phenomenon \\
T3 (Schema Control) & OFF & ON & Verify independent Schema compliance capability \\
\bottomrule
\end{tabular}
\end{table}

The three experimental conditions share identical System Prompt, User Prompt, tool definitions, and output parsing logic, with only the \texttt{tools} and \texttt{response\_format} API parameters adjusted. Through this design, observed behavioral differences can be attributed to joint constraints themselves rather than Prompt changes or task differences.

Each model runs 5 test rounds independently under each condition with fixed parameter settings:

\begin{itemize}
\item \texttt{temperature = 0.5}
\item \texttt{stream = true}
\item \texttt{max\_completion\_tokens = 4096}
\end{itemize}

According to the definitions in Section 3, Tool Invocation Rate (TIR) and Suppression Rate (SR) are recorded as primary evaluation metrics during experiments.

\subsection{Task Set Construction}

A common threat in tool-use evaluation is over-reliance on a single prompt instance.

To reduce this risk, evaluation tasks were sampled from representative production scenarios that require external information acquisition.

The task set covers five business-oriented categories:

\begin{itemize}
    \item Buyer Background Analysis
    \item Company Verification
    \item Market Intelligence Search
    \item Product Knowledge Retrieval
    \item Compliance and Risk Investigation
\end{itemize}

All tasks require information that is unavailable in model parameters alone and therefore necessitate at least one external tool invocation.

The same task set is executed across all T1/T2/T3 conditions, ensuring that observed differences originate from constraint configurations rather than task variation.

\begin{table}[ht]
    \centering
    \caption{Task Categories and Example Requirements}
    \label{tab:task_categories}
    \renewcommand\arraystretch{1.3}
    \begin{tabular}{ll}
        \toprule
        \textbf{Category} & \textbf{Example Requirement} \\
        \midrule
        Buyer Background & Analyze a buyer company \\
        Company Verification & Verify company legitimacy \\
        Market Search & Search target market demand \\
        Product Knowledge & Retrieve product specifications \\
        Compliance Check & Investigate regulatory risks \\
        \bottomrule
    \end{tabular}
\end{table}
For detailed task design, tool definitions, and schema specifications, see Appendix~\ref{app:test_design}.
\subsection{Model Selection}

To verify the universality of the Tool Suppression phenomenon, this paper selects seven model instances from different model families, parameter scales, and deployment methods, covering both closed-source baseline models and open-weight models.

\begin{table}[ht]
\centering
\caption{Evaluated Models}
\label{tab:models}
\small
\begin{tabular}{lcccc}
\toprule
Model & Parameters & Architecture & Deployment \\
\midrule
GPT-5.4-mini & --- & Proprietary & Cloud \\
Qwen3.6-35B-A3B & 35B & MoE (3B Active) & Local \\
Qwen3.5-122B-A10B & 122B & MoE (10B Active) & Local \\
GPT-OSS-20B & 20B & MoE (3.6B Active) & Local \\
Nemotron 3 Super & 120B & Hybrid MoE-Mamba & Local \\
Qwen3.5-397B-A17B & 397B & MoE (17B Active) & Cloud \\
Qwen3-VL-235B-Thinking & 235B & MoE + Vision-Language & Cloud \\
\bottomrule
\end{tabular}
\end{table}

Model selection follows these principles:

\begin{enumerate}
\item Covering different parameter scales (20B to 397B);
\item Covering different model architectures (MoE~\cite{garcia2025recency} and Hybrid architectures);
\item Covering local deployment and cloud deployment environments;
\item Including one closed-source commercial model as a reference baseline.
\end{enumerate}

This model matrix supports verification of RQ1 and RQ2 and provides a basis for subsequent cross-model comparisons. The purpose of this model selection is not to represent all open-weight LLMs, but to evaluate whether the observed behavior can be reproduced across multiple model families, parameter scales, deployment modes, and inference stacks.
The complete experimental environment configuration, including inference framework settings and parser configurations, is provided in Appendix~\ref{app:test_design}.

\subsection{Evaluation Protocol}

All experiments are executed through OpenAI-compatible API interfaces to ensure consistency in invocation methods across different models. Streaming generation mode is adopted during testing, with real-time parsing of tool call events and text content returned by models.

According to the definitions in Section 3, this paper adopts the following core metrics:

\begin{itemize}
    \item \textbf{Tool Invocation Rate (TIR)}: The proportion of sessions that contain at least one valid tool call.
    \item \textbf{Suppression Rate (SR)}: The relative reduction of tool invocation behavior under joint constraints.
    \item \textbf{JSON Compliance Rate (JCR)}: The proportion of responses that satisfy the required schema.
    \item \textbf{Average Tool Calls per Session (ATC)}: The average number of tool invocations generated within a successful session.
    \item \textbf{End-to-End Success Rate (ESR)}: The proportion of sessions that both invoke tools and produce valid structured outputs.
\end{itemize}

For each model, 5 test rounds are run independently under each experimental condition. Test tasks remain consistent with fixed random sampling parameters to reduce the impact of random fluctuations on results.

During streaming responses, the experimental framework automatically records the following information:

\begin{enumerate}
    \item Whether tool call events are generated;
    \item Number of tool calls;
    \item Tool call parameters;
    \item Final text response;
    \item JSON Schema compliance status.
\end{enumerate}

All experimental results are automatically collected and counted by the testing framework to avoid biases from manual recording.

For models supporting tool calling, if the response contains at least one valid tool call, it is recorded as a successful tool execution event; otherwise, it is recorded as no tool call. TIR and SR metrics are calculated based on this rule.

To ensure experimental reproducibility, all tests adopt unified Prompt templates, tool definition Schemas, and output format Schemas without special optimization for any individual model. Full details of the test scripts, detection logic, and schema definitions are documented in Appendix~\ref{app:test_design}.

\subsection{Parser-Level Validation}

A potential alternative explanation is that tool calls were generated but not captured correctly by the inference framework.

To exclude this possibility, raw streaming events were recorded during all experiments.

For each response, the framework simultaneously monitored:

\begin{enumerate}
    \item \texttt{tool\_calls} delta events;
    \item tool call argument streams;
    \item final assistant messages;
    \item parser outputs generated by the serving framework.
\end{enumerate}

A trial is considered a successful tool invocation only when a valid tool call event is observed in the raw stream.

Across all suppression cases reported in this paper, no \texttt{tool\_calls} delta events were observed.

This result indicates that the phenomenon cannot be explained by parser implementation failures or logging artifacts.

\subsection{Confounding Factor Controls}

To strengthen causal attribution, we systematically evaluated several alternative explanations.

\begin{table}[ht]
\centering
\caption{Confounding Factors and Tested Variants}
\label{tab:confounding_factors}
\renewcommand\arraystretch{1.3}
\begin{tabular}{ll}
\toprule
\textbf{Factor} & \textbf{Tested Variants} \\
\midrule
Framework & SGLang / vLLM \\
Schema Complexity & Simple / Medium / Production \\
Tool Choice & Optional / Required \\
Scale & 20B $\rightarrow$ 397B \\
Deployment & Local / Cloud \\
Parser Validation & Raw Stream Inspection \\
Fine-Tuning & Multiple SFT Variants \\
\bottomrule
\end{tabular}
\end{table}

\textbf{Inference Framework.} The same model was tested under both SGLang and vLLM while keeping prompts, schemas, tools, and model weights unchanged.

\textbf{Schema Complexity.} Experiments were conducted using simple, medium, and production-grade schemas to determine whether suppression is triggered only by highly complex output structures.

\textbf{Tool Enforcement Strategy.} Additional tests were performed using optional tool calls, explicit tool requirements, and \texttt{tool\_choice="required"} configurations.

\textbf{Model Scale.} The evaluated models span from 20B to 397B parameters.

\textbf{Deployment Environment.} Both local deployment and commercial cloud-hosted environments were included.

\textbf{Parser Reliability.} Raw streaming events were inspected to verify that missing tool calls were genuine behavioral outcomes rather than parser failures.

\textbf{Fine-Tuning Effects.} Multiple instruction-tuned variants trained on tool-use-oriented datasets were evaluated to determine whether conventional post-training mitigates the phenomenon.

Through the above control measures, this paper attributes Tool Suppression to the interaction effects between tool calling and structured output constraints to the greatest extent possible, rather than external factors such as model scale, inference framework, or deployment method.
\subsection{Root Cause Analysis: Grammar-Level Token Exclusion}

To determine whether the observed suppression behavior has a concrete implementation-level basis, we traced the inference stack from the \texttt{response\_format} API parameter to the token-generation layer. This investigation was conducted using SGLang 0.5.9, one of the two inference frameworks used in the main experiments.

The complete call chain is as follows:

\begin{center}
\begin{minipage}{0.85\textwidth}
\begin{lstlisting}[basicstyle=\ttfamily\small, frame=single, breaklines=true, numbers=none]
API: response_format={"type":"json_schema","json_schema":{...}}
    -> serving_chat.py:347: get_json_schema_constraint()
    -> grammar_manager.py:71: req.sampling_params.json_schema
    -> grammar_manager.py:89: xgrammar.compile_grammar()
    -> sampling_batch_info.py:197: update_regex_vocab_mask()
    -> sampling_batch_info.py:219: GrammarMatcher.fill_next_token_bitmask()
    -> bitmask_ops.py:74: apply_token_bitmask_inplace_triton()
    -> bitmask = ((packed_bitmask >> (0..31)) & 1) == 0
    -> logits[bitmask == 0] = -inf
    -> <tool_call> tokens unreachable
\end{lstlisting}
\end{minipage}
\end{center}

At the core of this mechanism is the vocabulary mask. The xgrammar engine compiles the JSON Schema into a finite-state machine (FSM). At each decoding step, the FSM queries which tokens are permitted given the current state. The inference framework then applies this mask to the logits: tokens with bit=0 are set to \(-\infty\), making them impossible to sample.

To verify that tool-call tokens are indeed masked, we directly inspected the FSM states for the JSON Schema used in the T2 condition. Table~\ref{tab:token_reachability} shows the reachability of key tokens across three representative FSM states.

\begin{table}[ht]
    \centering
    \caption{Token Reachability Under JSON Schema FSM}
    \label{tab:token_reachability}
    \renewcommand\arraystretch{1.3}
    \begin{tabular}{lccc}
        \toprule
        \textbf{Token} & \textbf{State 0 (Initial)} & \textbf{State 1 (Key)} & \textbf{State 2 (Value)} \\
        \midrule
        \texttt{\{} & \checkmark & \ding{55} & \ding{55} \\
        \texttt{"} & \ding{55} & \checkmark & \checkmark \\
        \texttt{<} & \ding{55} & \ding{55} & \ding{55} \\
        \texttt{<tool\_call>} & \ding{55} & \ding{55} & \ding{55} \\
        \texttt{</tool\_call>} & \ding{55} & \ding{55} & \ding{55} \\
        websearch & \ding{55} & \ding{55} & \ding{55} \\
        \bottomrule
    \end{tabular}
\end{table}

\texttt{<} (U+003C) is the first character of the Qwen model family's XML-style tool-call format. Since this character is not part of any valid JSON FSM state, the token is consistently masked across all generation states. As a result, the model cannot produce any token sequence starting with \texttt{<}, making tool-call tags unreachable.

This analysis establishes that Tool Suppression has a concrete implementation-level basis: when JSON Schema constraints are enforced through grammar-based constrained decoding, tool-call tokens are excluded from the decoding vocabulary. The phenomenon is not merely a behavioral tendency but a direct consequence of the decoding constraint mechanism.

The implication is that the suppression observed in Section 5 is at least partially attributable to this token-level exclusion. Whether models would have attempted to call tools in the absence of this mask cannot be determined from this analysis alone—the mask makes the question moot at the output layer.

% ===== 5. Empirical Findings =====
\section{Empirical Findings}

\subsection{RQ1: Reproducible Tool Suppression Across Tested Models}

RQ1 investigates whether the observed suppression behavior can be reproduced across different open-weight model families under identical task settings.

Table 5 summarizes the results obtained under the T1, T2, and T3 conditions. Across all evaluated open-weight models, tool execution behavior follows a highly consistent pattern.

Under T1 (Tools ON, Schema OFF), all evaluated models successfully invoked tools when external information was required. Tool Invocation Rate (TIR) reached 100\% across all tested sessions.

Under T2 (Tools ON, Schema ON), tool execution behavior disappeared completely. No valid tool call events were observed in any evaluated session despite the continued presence of tool definitions and unchanged task requirements.

Under T3 (Tools OFF, Schema ON), models generally maintained high schema compliance rates, demonstrating that structured output generation capability itself remained functional.

This result indicates that both capabilities exist independently:

\begin{enumerate}
    \item Tool execution capability is available under T1.
    \item Schema compliance capability is available under T3.
    \item The combination of both constraints under T2 introduces a systematic behavioral change.
\end{enumerate}

Importantly, the observed pattern is reproduced across multiple model families, parameter scales, deployment environments, and inference stacks. While the evaluated closed-source reference model (GPT-5.4-mini) maintained stable tool execution behavior under all conditions, all tested open-weight models exhibited complete suppression under T2.

Therefore, within the evaluated model set, tool suppression appears to be a reproducible phenomenon rather than an isolated model defect. The test design and query diversity are further detailed in Appendix~\ref{app:test_design}.

\begin{table}[ht]
\centering
\caption{Results under T1/T2/T3 Conditions}
\label{tab:full_results}
\small
\renewcommand\arraystretch{1.3}
\begin{tabular}{lcccc}
\toprule
Model & T1 TIR & T2 TIR & T3 JC & SR \\
\midrule
GPT-5.4-mini & 100\% & 100\% & 100\% & 0\% \\
Qwen3.6-35B-A3B (SGLang) & 100\% & 0\% & 100\% & 100\% \\
Qwen3.6-35B-A3B (vLLM) & 100\% & 0\% & 80\% & 100\% \\
Qwen3.5-122B-A10B & 100\% & 0\% & 100\% & 100\% \\
GPT-OSS-20B & 100\% & 0\% & 100\% & 100\% \\
Nemotron 3 Super & 100\% & 0\% & 100\% & 100\% \\
Qwen3.5-397B-A17B & 100\% & 0\% & 100\% & 100\% \\
Qwen3-VL-235B-Thinking & 100\% & 0\% & 100\% & 100\% \\
\bottomrule
\end{tabular}
\end{table}

\subsection{RQ2: Evidence for Constraint Interaction}

A key question is whether the observed behavior originates from a deficiency in tool-use capability or schema-following capability individually.

The experimental results do not support either explanation.

Under T1, models successfully invoke tools when structured output constraints are absent. Under T3, models successfully generate schema-compliant outputs when tool execution requirements are absent.

The failure emerges only under T2, where both constraints coexist.

This observation suggests that the suppression behavior is associated with the interaction between Tool Calling and Structured Output constraints rather than with either capability in isolation.

The result is particularly notable because the same prompts, task requirements, tool definitions, and schemas are used across all conditions. The only difference lies in the simultaneous activation of both constraint mechanisms.

Therefore, the evidence supports the interpretation that suppression emerges from a joint-constraint setting rather than from a general inability to use tools or follow schemas.

\subsection{Schema Complexity Ablation}

One possible explanation is that suppression is triggered only when output schemas become sufficiently complex.

To evaluate this possibility, additional experiments were conducted using three schema complexity levels:

\begin{itemize}
    \item \textbf{Simple Schema}: minimal output structure with only a small number of required fields.
    \item \textbf{Medium Schema}: intermediate business-oriented structures containing multiple nested fields.
    \item \textbf{Production Schema}: full production response formats containing numerous required objects and validation constraints.
\end{itemize}

\begin{table}[ht]
    \centering
    \caption{Schema Complexity Ablation Results}
    \label{tab:schema_complexity}
    \renewcommand\arraystretch{1.3}
    \begin{tabular}{lcc}
        \toprule
        \textbf{Schema} & \textbf{Fields} & \textbf{T2 TIR} \\
        \midrule
        Simple & 1--3 & 0\% \\
        Medium & 5--10 & 0\% \\
        Production & 20+ & 0\% \\
        \bottomrule
    \end{tabular}
\end{table}

Across all three schema categories, suppression behavior remained unchanged. Tool Invocation Rate under T2 remained at 0\% regardless of schema complexity.

No threshold effect was observed.

These results suggest that suppression is not exclusively associated with highly complex production schemas. Instead, even relatively lightweight schema constraints appear sufficient to trigger the observed behavior.

Therefore, schema complexity may influence formatting difficulty, but it does not appear to be the primary factor responsible for tool execution suppression.

\subsection{Tool Enforcement Ablation}

A second alternative explanation is that suppressed models simply fail to recognize that tool use is required.

To investigate this possibility, multiple tool invocation strategies were evaluated:

\begin{itemize}
    \item Optional tool use;
    \item Explicit system-prompt instructions requiring tool execution;
    \item Explicit user-prompt instructions requiring tool execution;
    \item API-level \texttt{tool\_choice="required"}.
\end{itemize}

\begin{table}[ht]
    \centering
    \caption{Tool Enforcement Ablation Results}
    \label{tab:tool_enforcement}
    \renewcommand\arraystretch{1.3}
    \begin{tabular}{lc}
        \toprule
        \textbf{Setting} & \textbf{T2 TIR} \\
        \midrule
        Optional & 0\% \\
        User Prompt Required & 0\% \\
        System Prompt Required & 0\% \\
        \texttt{tool\_choice="required"} & 0\% \\
        \bottomrule
    \end{tabular}
\end{table}

Results remained consistent across all configurations.

Even when tool execution was explicitly mandated through API-level enforcement mechanisms, no tool call events were generated under T2.

This finding is important because it suggests that suppression cannot be fully explained by incorrect task interpretation or weak prompting.

Instead, the observed behavior persists even when tool invocation requirements are made explicit at multiple control layers.

Consequently, the results provide additional evidence that the failure occurs after tool necessity has already been established.

\subsection{Fine-Tuning Does Not Eliminate Suppression}

To investigate whether conventional post-training techniques~\cite{xu2023wizardlm} can mitigate suppression, we evaluated multiple instruction-tuned variants trained using tool-use-oriented datasets and production agent trajectories.

The evaluated variants include:

\begin{itemize}
    \item Tool-use supervised fine-tuning datasets;
    \item Schema-aware instruction tuning datasets;
    \item Production business-agent trajectories;
    \item Reinforcement-learning-enhanced variants.
\end{itemize}

\begin{table}[ht]
    \centering
    \caption{Fine-Tuning Ablation Results}
    \label{tab:finetuning}
    \renewcommand\arraystretch{1.3}
    \begin{tabular}{lcc}
        \toprule
        \textbf{Model Variant} & \textbf{Samples} & \textbf{T2 TIR} \\
        \midrule
        Base & -- & 0\% \\
        Tool Mandatory & 200 & 0\% \\
        Schema Injection & 200 & 0\% \\
        GRPO & 200 & 0\% \\
        Large SFT & 6000 & 0\% \\
        \bottomrule
    \end{tabular}
\end{table}

Despite improvements in general instruction following and tool-use behavior under T1 conditions, suppression remained unchanged under T2.

Across all evaluated variants, Tool Invocation Rate under joint constraints remained at or near zero.

These results suggest that the phenomenon is not easily removed through conventional post-training approaches alone.

While the evaluated fine-tuning configurations do not exhaust all possible alignment strategies, the evidence indicates that suppression may be more deeply connected to action-selection behavior than to basic tool-use capability.

\subsection{RQ3: Framework Independence}

To determine whether suppression originates from inference infrastructure rather than model behavior, identical experiments were conducted using both SGLang and vLLM.

All prompts, schemas, tool definitions, model weights, and hardware configurations were held constant.

The two frameworks exhibited minor differences in schema compliance rates, but their tool execution behavior remained identical.

Under T1, tool execution succeeded consistently.

Under T2, tool execution disappeared completely.

Because the same behavioral pattern appears across independent serving stacks, the results do not support the hypothesis that suppression originates from framework-specific parser implementations or scheduling mechanisms.

Instead, the evidence suggests that the phenomenon is primarily associated with model-level behavior.
\subsection{Finding: Token-Level Exclusion Under Grammar-Constrained Decoding}

The root cause analysis presented in Section 4.6 reveals a concrete mechanism underlying the observed suppression behavior.

When \texttt{response\_format} is enabled, the inference framework compiles the JSON Schema into a finite-state machine and applies a vocabulary mask at each decoding step. Tokens that do not conform to the current FSM state are assigned logits of \(-\infty\), making them impossible to sample.

For the Qwen model family used in this study, tool calls are formatted as XML-style tags beginning with \texttt{<tool\_call>}. The \texttt{<} character (U+003C) is not part of any valid JSON token sequence. Consequently, in all FSM states—whether the model is at the start of the response, in a key position, or in a value position—\texttt{<} and any token beginning with \texttt{<} are masked.

This finding establishes that Tool Suppression has a concrete implementation-level basis: under grammar-constrained decoding, tool-call tokens become unreachable by design. The absence of tool calls under T2 is therefore not simply a model preference or behavioral tendency, but a direct consequence of the decoding constraint mechanism.

Importantly, this does not preclude the existence of additional model-level behavioral factors. However, it confirms that at least one concrete root cause exists at the inference-stack level.

\subsection{Threat Elimination Summary}

The experiments collectively eliminate several common alternative explanations.

Suppression is reproduced across:

\begin{itemize}
    \item Multiple model families;
    \item Multiple parameter scales;
    \item Multiple deployment environments;
    \item Multiple inference frameworks;
    \item Multiple schema complexity levels;
    \item Multiple tool invocation strategies;
    \item Multiple fine-tuning configurations.
\end{itemize}

In addition, parser-level inspection confirms that missing tool calls are genuine behavioral outcomes rather than logging failures.

Taken together, the evidence consistently points toward a behavioral interaction between tool execution and structured output constraints.

While the precise mechanism remains an open question, the observed phenomenon cannot be readily explained by model size, framework implementation, deployment configuration, schema complexity, or prompt-level enforcement alone.

% ===== 6. Behavioral Analysis =====

\section{Mechanism Analysis and Behavioral Interpretation}

The empirical findings in Section 5 establish that Tool Suppression is reproducible across the evaluated model set. The root cause analysis in Section 4.6 further identifies a concrete implementation-level mechanism: grammar-constrained decoding excludes tool-call tokens from the vocabulary. This section examines the relationship between this token-level exclusion and the observed behavioral patterns.

\subsection{Grammar-Level Tool Exclusion}

The inference-stack tracing in Section 4.6 reveals that JSON Schema constraints are enforced through grammar-based token masking. This mechanism has three key characteristics:

First, the mask is applied at every decoding step. The FSM state determines which tokens are permissible, and any token not permitted by the current state receives a logit of \(-\infty\).

Second, the mask is universal within the constrained generation context. For Qwen-family models that use XML-style \texttt{<tool\_call>} tags, the \texttt{<} character is never permitted in any JSON FSM state. Therefore, tool-call tokens are unreachable throughout the entire generation process.

Third, the mask operates independently of model weights. Even if the model's internal logit distribution assigns high probability to \texttt{<tool\_call>}, the mask sets that logit to \(-\infty\) before sampling. The model's preference is overridden by the decoding constraint.

This mechanism explains why conventional post-training approaches—including the SFT and GRPO variants evaluated in Section 5.5—failed to eliminate suppression. These methods modify model weights, which affect the logit distribution before the mask is applied. However, the mask operates after the model produces logits and before sampling. There is no gradient path from the mask back to the weights, so weight-level optimization cannot overcome token-level exclusion.

\begin{figure}[ht]
    \centering
    \includegraphics[width=0.85\textwidth]{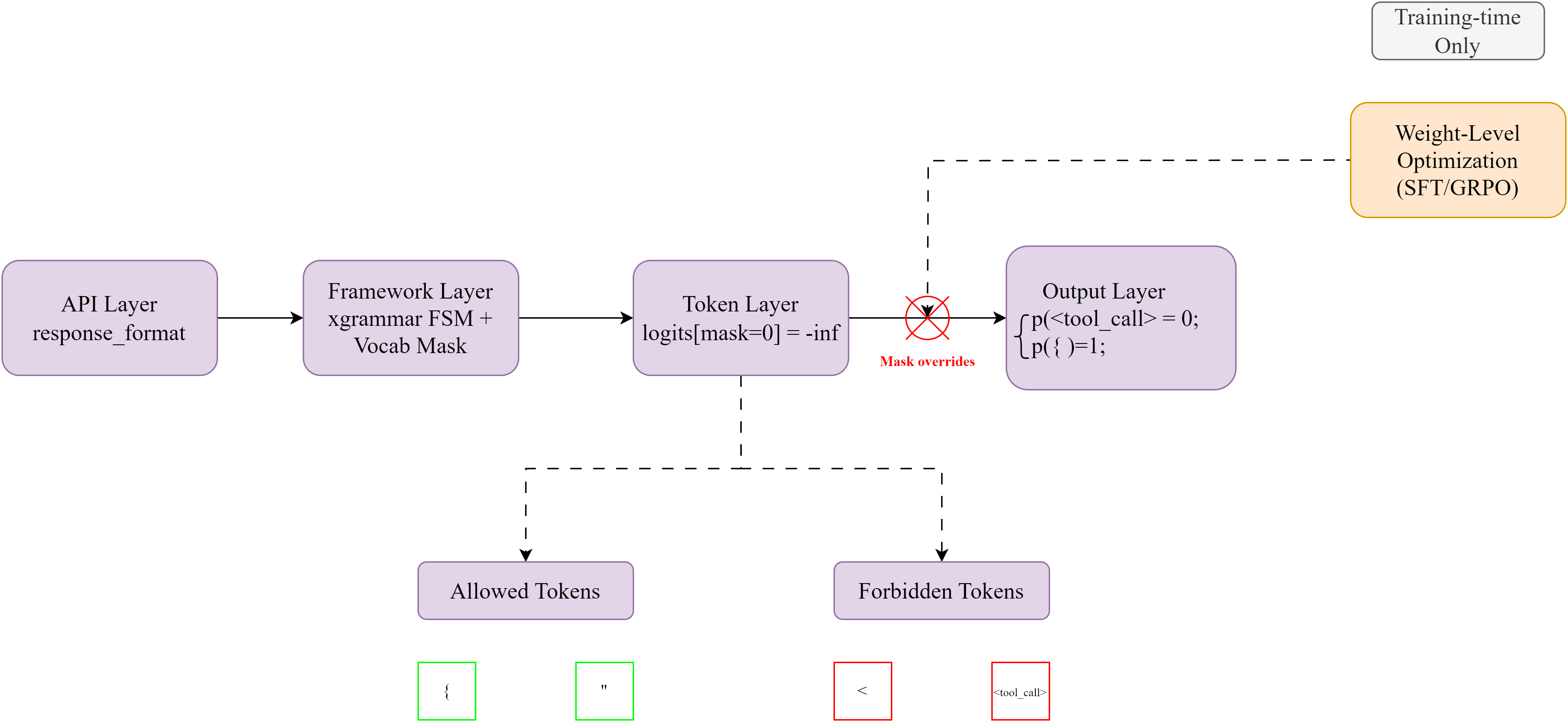}
    \caption{Constraint Tax Mechanism: Token-Level Masking Overrides Model Preferences}
    \label{fig:mechanism}
\end{figure}

Figure~\ref{fig:mechanism} illustrates this architecture. The API-layer constraint triggers grammar compilation, which produces a vocabulary mask. This mask is applied to the model's logits before sampling, rendering non-JSON tokens—including tool-call tags—unreachable. Weight-level optimization (SFT/DPO/GRPO) modifies the logits but cannot bypass the mask.

\subsection{Relationship to Constraint Priority Inversion}

The grammar-level exclusion mechanism described above provides a concrete explanation for why tool-call tokens are absent under T2. However, it does not fully explain all observed behavioral patterns.

In particular, the diagnostic experiment described in Section 6.1 of the original analysis revealed that models sometimes produced outputs containing \texttt{need\_search: true} while generating no corresponding tool calls. This pattern—classified as TS-C in the taxonomy—indicates that models can explicitly acknowledge tool requirements even when tool execution is impossible.

This observation suggests that the behavioral phenomenon may involve two layers:

\begin{itemize}
    \item \textbf{Decoding layer:} Grammar-constrained decoding makes tool-call tokens unreachable. This explains why no tool calls are generated.
    \item \textbf{Model layer:} Models may still recognize tool requirements and express this recognition in natural language or structured fields. This explains why some models produce TS-C patterns.
\end{itemize}

The Constraint Priority Inversion (CPI) hypothesis, introduced in Section 6.3 of this paper, provides a behavioral interpretation of the model-layer phenomenon. CPI proposes that when Tool Calling and Structured Output constraints coexist, models may prioritize schema satisfaction in their generation strategy. This is consistent with the observation that models continue to produce schema-compliant responses while tool calls are absent.

However, the CPI hypothesis is not the only possible interpretation. The model's expressed tool awareness in TS-C patterns may simply reflect its internal representation of task requirements, which is independent of the decoding constraints that prevent execution. Whether CPI represents a distinct behavioral mechanism or an artifact of the model's training distribution remains an open question.

At present, we interpret the evidence as follows:

\begin{enumerate}
    \item The \textbf{primary cause} of missing tool calls under T2 is grammar-level token exclusion at the decoding layer.
    \item The \textbf{behavioral patterns} observed across models—including TS-C's expressed tool awareness—are consistent with CPI as a behavioral hypothesis, but additional research is needed to determine whether CPI reflects a distinct internal mechanism.
\end{enumerate}

This two-layer interpretation reconciles the implementation-level evidence with the behavioral observations, while maintaining appropriate caution regarding causal claims about internal model mechanisms.

% ===== 7. Mitigation Through Transparent Two-Pass Execution =====
\section{Mitigation Through Transparent Two-Pass Execution}

\subsection{Design Motivation}

The findings presented in Sections 5 and 6 indicate that suppression emerges when Tool Calling and Structured Output constraints are activated simultaneously. The root cause analysis in Section 4.6 further reveals that JSON Schema constraints are enforced through grammar-based token masking, which renders tool-call tokens unreachable during decoding.

This observation suggests a straightforward engineering question:

Can the two constraints be decoupled such that tool execution occurs before grammar-based token masking is applied?

The proposed mitigation strategy is motivated by a simple principle:

Execute tools first, then enforce structured output constraints.

By separating the two phases, the model can invoke tools freely in the first pass without any grammar constraints. In the second pass, once tool results are collected and incorporated into the context, the model is asked to generate the final schema-compliant response under grammar constraints.

This design is informed by the mechanism identified in Section 4.6: the root cause of suppression is that grammar-constrained decoding makes tool-call tokens unreachable. The mitigation strategy avoids this problem by ensuring that tool execution is completed before the grammar constraint is activated.

This design does not modify model weights, training data, inference kernels, or serving frameworks. Therefore, it can be deployed directly within existing production Agent systems without requiring model retraining.

The goal of the approach is not to improve tool-use capability itself, but to avoid the token-level exclusion caused by simultaneous constraint activation.

\subsection{Transparent Two-Pass Execution}

The proposed mitigation strategy consists of two sequential inference stages.

\textbf{Pass 1: Tool Execution Phase}

\begin{itemize}
    \item Tools = ON
    \item Schema = OFF
\end{itemize}

The model is allowed to freely invoke tools and perform multi-step information acquisition without structured output constraints.

All tool outputs are collected and preserved.

\textbf{Pass 2: Structured Output Phase}

\begin{itemize}
    \item Tools = OFF
    \item Schema = ON
\end{itemize}

The collected tool results are injected into the context, and the model is re-invoked to generate the final schema-compliant response.

The complete execution flow is:

\underline{Execution Flow:}

User Request $\rightarrow$ Pass 1 Tool Execution $\rightarrow$ Tool Results Collection $\rightarrow$ Pass 2 Structured Output Generation $\rightarrow$ Final Response

This architecture changes only the order in which constraints are applied.

No modifications are made to model parameters, decoding algorithms, or training procedures.

For this reason, the approach can be viewed as an inference-time mitigation strategy rather than a model-level solution.

Figure 2 illustrates the Transparent Two-Pass Execution architecture.

\begin{figure}[ht]
    \centering
    \includegraphics[width=0.8\textwidth]{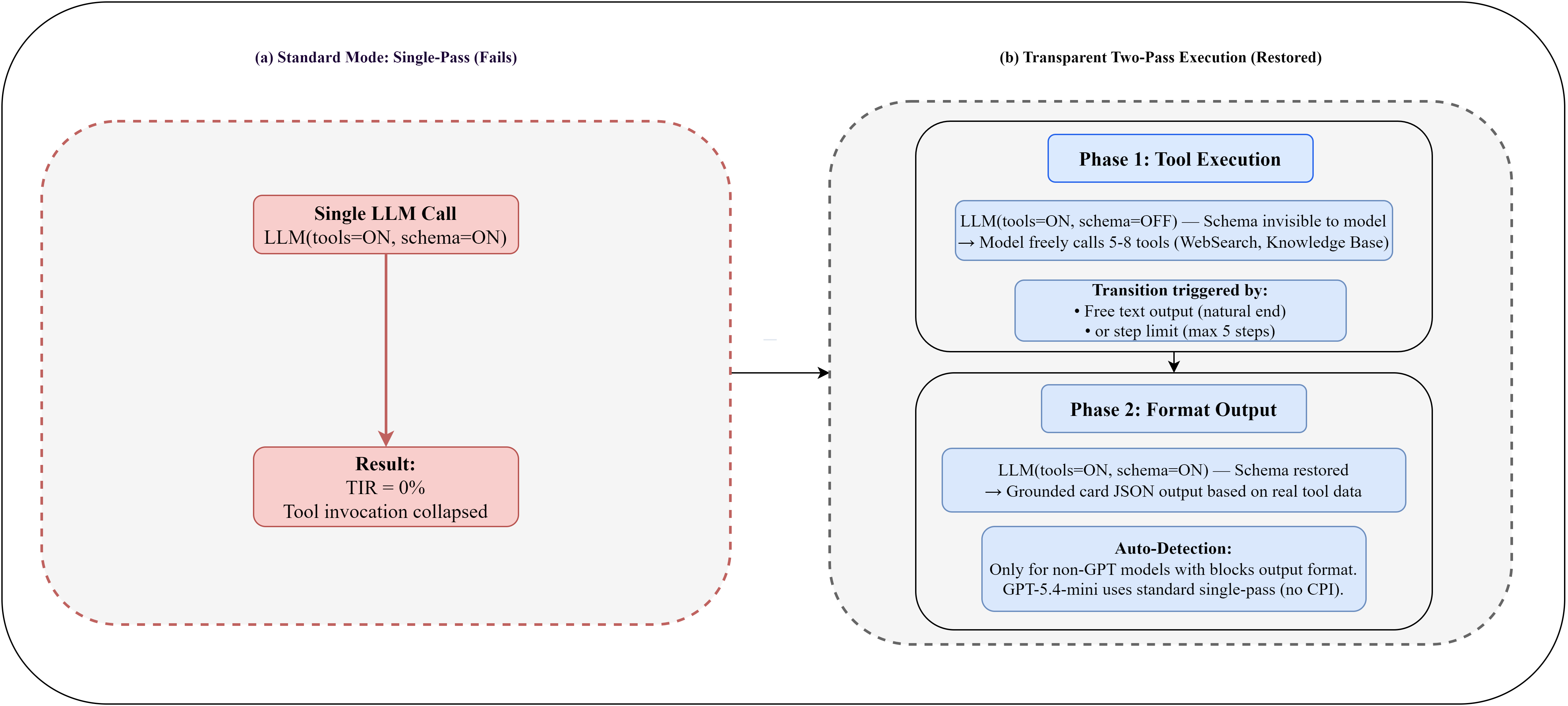}
    \caption{Transparent Two-Pass Execution Architecture}
    \label{fig:two_pass}
\end{figure}

\subsection{Experimental Evaluation}

To evaluate the effectiveness of Transparent Two-Pass Execution, the strategy was deployed within the production Agent pipeline and tested on the same task categories used in previous experiments.

\begin{table}[ht]
    \centering
    \caption{Mitigation Effectiveness Before and After Two-Pass Execution}
    \label{tab:mitigation_results}
    \renewcommand\arraystretch{1.3}
    \begin{tabular}{lcc}
        \toprule
        \textbf{Metric} & \textbf{Before (T2)} & \textbf{After (Two-Pass)} \\
        \midrule
        Tool Invocation Rate & 0\% & 100\% \\
        JSON Compliance Rate & 100\% & 100\% \\
        End-to-End Success Rate & 0\% & 100\% \\
        Avg Tool Calls / Session & 0 & 5--8 \\
        \bottomrule
    \end{tabular}
\end{table}

Tool Invocation Rate increased from 0\% to 100\%.

JSON Compliance Rate remained unchanged.

End-to-End Success Rate improved from complete task failure to successful completion.

Most importantly, restored tool execution enabled the system to utilize real external information rather than generating responses based solely on parametric knowledge.

As a result, behaviors previously observed under suppression conditions, including simulated retrieval and tool-free hallucination, were substantially reduced.

These results demonstrate that separating tool execution from structured output generation can effectively restore agent functionality without sacrificing schema compliance.

\subsection{Cost and Latency Analysis}

Although Transparent Two-Pass Execution successfully restores tool execution behavior, the approach introduces additional inference overhead.

Compared with the original single-pass architecture, the mitigation requires an additional model invocation.

Consequently, latency increases approximately by one additional inference round plus tool execution time.

Token consumption also increases because the second pass receives both the original request and the collected tool outputs as context.

\begin{table}[ht]
    \centering
    \caption{Cost and Latency Comparison}
    \label{tab:cost_latency}
    \renewcommand\arraystretch{1.3}
    \begin{tabular}{lcc}
        \toprule
        \textbf{Metric} & \textbf{Single Pass (T2)} & \textbf{Two Pass} \\
        \midrule
        LLM Rounds & 1 & 2 \\
        Tool Calls & 0 & 5--8 \\
        Success Rate & 0\% & 100\% \\
        \bottomrule
    \end{tabular}
\end{table}

However, in the evaluated production workflow, the additional cost was considered acceptable because the baseline system under suppression conditions could not successfully complete the target tasks.

Therefore, the mitigation exchanges additional latency and token usage for functional correctness.

This tradeoff is often favorable in enterprise Agent scenarios where task completion is more important than minimizing inference cost.

\subsection{Failure Cases and Limitations}

Transparent Two-Pass Execution is an engineering mitigation rather than a complete solution.

Several limitations remain.

First, the approach introduces additional latency because two separate model invocations are required.

Second, token consumption increases due to repeated context transmission between passes.

Third, the strategy relies on external orchestration logic and therefore does not eliminate the underlying behavioral mechanism responsible for suppression.

Finally, the current evaluation focuses primarily on tool-calling scenarios. Whether similar benefits extend to other agent architectures involving workflow execution, MCP ecosystems, or multi-agent collaboration remains an open question.

Therefore, the proposed approach should be viewed as a practical production workaround rather than a definitive solution to the suppression problem.

% ===== 8. Discussion =====
\section{Discussion}

\subsection{What Constraint Tax Means for Agent Evaluation}

An important implication of this study concerns how Agent capabilities are evaluated.

Current evaluation frameworks frequently assess tool use and structured output generation as independent capabilities.

Tool-use evaluations typically measure:

\begin{itemize}
    \item Tool selection accuracy;
    \item Parameter generation quality;
    \item Task completion performance.
\end{itemize}

Structured-output evaluations typically measure:

\begin{itemize}
    \item Schema compliance;
    \item Field completeness;
    \item Formatting correctness.
\end{itemize}

The results of this study suggest that strong performance on these dimensions individually does not necessarily imply reliable behavior when both constraints are activated simultaneously.

In the evaluated tasks, models were capable of tool execution under T1 and schema compliance under T3, yet failed to execute tools under T2.

This observation indicates that joint-constraint scenarios may expose behaviors that are not visible when capabilities are evaluated independently.

Therefore, future Agent evaluation suites may benefit from including explicit Tool Calling + Structured Output conditions as a separate evaluation setting rather than assuming that independent capability measurements are sufficient.

\subsection{Implications for Agent System Design}

The observed suppression behavior, now traced to grammar-based token masking, highlights an important engineering consideration for production Agent systems.

In many practical deployments, tool execution and structured outputs are treated as complementary capabilities that can be activated simultaneously without additional coordination. However, the results presented in this study suggest that the interaction between these capabilities may be mediated by the inference framework's constraint enforcement mechanism.

From a systems perspective, reliability under joint constraints may require explicit orchestration strategies rather than relying entirely on model-level behavior. The success of Transparent Two-Pass Execution demonstrates that separating tool execution from grammar-constrained generation is one effective approach.

More broadly, the findings suggest that Agent architecture design must consider not only model intelligence but also the interaction between model behavior, inference framework constraints, and token-level decoding mechanisms.

Consequently, future Agent platforms may need to evaluate not only model accuracy but also how inference-level constraints—such as grammar masks—interact with tool execution and structured output requirements.

\subsection{Interpreting the Open-Weight Results}

One noteworthy observation from the evaluated model set is the behavioral difference between the tested open-weight models and the closed-source reference model included in this study.

All evaluated open-weight models exhibited complete suppression under T2, whereas GPT-5.4-mini maintained stable tool execution behavior.

The root cause analysis in Section 4.6 provides a plausible explanation for this difference: the suppression observed in open-weight models can be traced to grammar-based token masking in the inference framework. GPT-5.4-mini's behavior may differ because its implementation uses a different mechanism for enforcing structured outputs—potentially through instruction-level guidance rather than decoding-level grammar constraints, or through a different grammar implementation that permits interleaved tool calls.

However, caution is required when interpreting this difference. The current study evaluates only a limited number of models and does not provide visibility into proprietary training procedures, alignment pipelines, or deployment architectures. Several explanations remain plausible, including differences in:

\begin{itemize}
    \item Structured-output implementation (grammar vs. instruction-based);
    \item Tokenizer design and tool-call format;
    \item Post-training objectives and data distributions;
    \item Inference-time orchestration strategies.
\end{itemize}

Consequently, the results should be interpreted as evidence of behavioral divergence within the evaluated models, with a concrete root cause identified for the open-weight cases, rather than as evidence of a universal distinction between open-weight and closed-source systems.
\subsection{Limitations}

Several limitations should be considered when interpreting the findings of this study.

First, the evaluation focuses on a finite set of open-weight models and one closed-source reference model. The results therefore support conclusions regarding the evaluated models rather than all large language models.

Second, although multiple task categories were included, the benchmark remains substantially smaller than large-scale academic evaluation suites.

Third, the CPI framework is currently a behavioral hypothesis derived from observed outcomes rather than a verified internal mechanism. The present study identifies consistent external behaviors but does not directly observe model internals.

Fourth, the mitigation evaluation focuses primarily on production tool-calling workflows. Whether similar effects appear in broader agent architectures such as MCP ecosystems, multi-agent systems, workflow agents, or computer-use agents remains an open question.

Finally, the study emphasizes practical reproducibility and engineering diagnosis rather than theoretical guarantees. Future work will be required to determine the precise mechanisms responsible for the observed behavior.

% ===== 9. Future Work =====
\section{Future Research Directions}

\subsection{Toward Standardized Joint-Constraint Benchmarks}

The experiments presented in this study suggest that joint-constraint scenarios deserve explicit evaluation rather than being treated as a simple combination of existing capabilities.

Current benchmarks typically evaluate tool use, structured output generation, and task completion separately. However, the observed suppression behavior emerges specifically when these capabilities must operate simultaneously.

Future benchmark development may therefore benefit from introducing dedicated joint-constraint evaluation tracks that include:

\begin{itemize}
    \item Tool Calling + Structured Output tasks;
    \item Multiple schema complexity levels;
    \item Multiple tool-call formats (XML, function-call, etc.);
    \item Cross-framework evaluation protocols;
    \item Reliability-oriented metrics beyond task accuracy.
\end{itemize}

Such benchmarks would help quantify the prevalence of suppression behaviors and provide a common basis for comparing mitigation strategies across different model families and agent architectures.

\subsection{Grammar-Aware Tool Calling and Decoding}

The root cause analysis in this paper identifies grammar-based token masking as a concrete mechanism underlying Tool Suppression. This finding suggests several directions for future work on inference frameworks.

First, inference frameworks could be extended to support grammar specifications that explicitly permit tool-call tokens alongside JSON Schema constraints. This would require modifying the FSM to allow interleaved tool-call sequences without sacrificing format compliance.

Second, frameworks could provide more transparent feedback to developers when tool-call tokens are masked. Currently, the masking is silent—models generate schema-compliant outputs without indicating that tool calls were prevented. Debugging this behavior currently requires source-code tracing.

Third, the interaction between grammar constraints and different tool-call formats (XML vs. function-call vs. MCP) merits further investigation. Different formats may be more or less compatible with JSON Schema FSMs.

\subsection{Understanding the Relationship Between Token Masking and Model Behavior}

The mechanism identified in this paper—token-level exclusion—does not fully explain all observed behavioral patterns, particularly TS-C (Intent Without Action).

Future research could investigate whether models' expressed tool awareness in TS-C patterns reflects:

\begin{itemize}
    \item Internal recognition of tool requirements that is independent of decoding constraints;
    \item A behavioral strategy learned from training data where tool intent is expressed without execution;
    \item An artifact of the interaction between the model's output distribution and the applied mask.
\end{itemize}

Methods for this investigation could include activation analysis, controlled training experiments, and systematic variation of grammar constraints.

\subsection{Beyond Tool Calling}

This study focuses specifically on tool-calling workflows because they represent a common production deployment pattern and provide a clear observable signal for behavioral analysis.

However, the token-level exclusion mechanism identified in this paper is not inherently limited to tool calling. Any agent behavior that requires generating token sequences outside the JSON grammar—such as workflow control tokens, multi-agent communication markers, or computer-use action sequences—may be similarly affected.

Future research should investigate whether similar exclusion effects occur in:

\begin{itemize}
    \item Workflow execution with structured control flows;
    \item Multi-agent protocols with specialized token formats;
    \item MCP-based tool ecosystems with custom action tags;
    \item Computer-use agents with action sequences.
\end{itemize}

If comparable patterns are observed across different agent architectures, the phenomenon described in this paper may represent a broader class of constraint-driven behavioral exclusions rather than a tool-calling-specific issue.

\section{Conclusion}

This paper investigates the behavior of open-weight large language models under joint Tool Calling and Structured Output constraints and reports a reproducible failure pattern observed in a production Agent system.

Across the evaluated model set, tool execution and schema compliance remain functional when assessed independently. However, when both constraints are activated simultaneously, tool execution behavior disappears despite the continued presence of tool definitions, unchanged task requirements, and successful schema generation.

To characterize this phenomenon, we introduce the term Tool Suppression and provide evidence from controlled experiments covering multiple model families, parameter scales, deployment environments, inference frameworks, schema configurations, tool invocation strategies, and post-training variants.

The results suggest that the observed behavior cannot be readily explained by model size, framework implementation, deployment configuration, schema complexity, prompt-level enforcement, or conventional instruction tuning alone.

Through systematic tracing of the inference stack, we identify one concrete root cause: JSON Schema constraints are compiled into grammar-based token masks that render tool-call tokens unreachable during decoding. This mechanism operates at the inference-framework layer, independent of model weights, and explains why weight-level optimization approaches such as SFT and GRPO do not eliminate suppression.

To interpret the observed behavioral patterns, we formulate the Constraint Priority Inversion (CPI) hypothesis. CPI proposes that, under joint constraints, schema satisfaction may dominate action selection behavior. The current study presents CPI as a behavioral hypothesis consistent with the observed evidence rather than as a verified internal mechanism.

At the engineering level, this paper proposes Transparent Two-Pass Execution, an inference-time mitigation strategy that separates tool execution from schema-constrained response generation. Experimental results demonstrate that this approach can restore tool invocation behavior while preserving structured output guarantees, without requiring modifications to model weights or training procedures.

More broadly, the findings suggest that evaluating tool use and structured output generation independently may not fully capture the reliability of modern Agent systems operating under multiple simultaneous constraints. The interaction between inference framework constraints and model behavior—particularly token-level exclusion mechanisms—deserves further investigation from both the model and system perspectives.

The primary contributions of this work are:

\begin{itemize}
    \item Identification and characterization of a reproducible Tool Suppression phenomenon under joint Tool Calling and Structured Output constraints;
    \item Systematic evaluation of alternative explanations across models, frameworks, schemas, tool invocation strategies, and post-training configurations;
    \item Localization of a concrete root cause to grammar-based token masking in constrained decoding;
    \item Formulation of the Constraint Priority Inversion (CPI) hypothesis as a behavioral interpretation of the observed suppression pattern;
    \item Proposal and validation of Transparent Two-Pass Execution as a practical mitigation strategy for production Agent systems.
\end{itemize}

Although the phenomenon reported in this paper was discovered through a specific production deployment, the underlying challenge is increasingly relevant as modern Agent systems combine tool use, structured outputs, workflow execution, and external orchestration within a single interaction loop.

As Agent architectures continue to evolve, understanding how multiple constraints—both behavioral and implementation-level—interact during action selection may become as important as improving reasoning capability itself.

We hope this study contributes to a deeper understanding of behavioral reliability in Agent systems and encourages future research on the interaction between tool execution, structured generation, decoding constraints, and action selection under competing objectives.

% ===== References =====
\bibliographystyle{unsrt}
\bibliography{references}

\appendix
\section{Test Case Design and Tool/Schema Definitions}
\label{app:test_design}
This appendix provides detailed documentation of the test case design, tool definitions, schema specifications, and experimental configurations used in the main experiments. The materials are provided to support reproducibility and to clarify the exact conditions under which Tool Suppression was observed.

\subsection{Controlled Experimental Design}

Each evaluated model was tested independently under three conditions. Table~\ref{tab:app_conditions} summarizes the experimental configuration.

\begin{table}[ht]
\centering
\caption{T1/T2/T3 Experimental Conditions}
\label{tab:app_conditions}
\renewcommand\arraystretch{1.3}
\begin{tabular}{lccc}
\toprule
Condition & \texttt{tools} & \texttt{response\_format} & Purpose \\
\midrule
T1 (Baseline) & ON & OFF & Measure baseline tool calling capability \\
T2 (Joint Constraint) & ON & ON & \textbf{Detect Tool Suppression} \\
T3 (Schema Control) & OFF & ON & Verify independent schema compliance \\
\bottomrule
\end{tabular}
\end{table}

\subsection{Standardized Test Script Protocol}

All cross-model tests used a unified Python testing framework. The core test scripts are organized as follows:

\begin{itemize}
    \item \texttt{test\_constraint\_tax\_lora.py} — Qwen3.6-35B-A3B LoRA fine-tuning variants
    \item \texttt{test\_cloud\_models.py} — GPT-5.4-mini, Qwen3.5-397B-A17B, Qwen3-VL-235B
    \item \texttt{test\_122b\_tool\_response\_format.py} — Qwen3.5-122B-A10B
    \item \texttt{test\_gptoss\_tool\_rfmt.py} — GPT-OSS-20B
    \item \texttt{test\_vllm\_qwen35b.py} — Qwen3.6-35B-A3B via vLLM
    \item \texttt{test\_nemotron\_tool\_rfmt.py} — NVIDIA Nemotron 3 Super
    \item \texttt{test\_b1\_two\_pass.py} — Two-Pass end-to-end validation
\end{itemize}

All scripts share the following standard parameters:

\begin{lstlisting}[basicstyle=\ttfamily\small, frame=single, breaklines=true, numbers=none]
ROUNDS  = 5          # Independent test rounds per condition
temperature = 0.5
stream = True
max_completion_tokens = 4096
\end{lstlisting}

The unified detection logic across all scripts uses dual validation:

\begin{lstlisting}[basicstyle=\ttfamily\small, frame=single, breaklines=true, numbers=none]
async for chunk in resp:
    if d.tool_calls:      # API-level: parse streaming delta tool_calls
        for tc in d.tool_calls:
            tcs.append({"idx": tc.index, "name": tc.function.name, ...})
    if d.content:         # Text-level: collect content for JSON compliance check
        content += d.content
\end{lstlisting}

\subsection{Task Set and Diversity}

\subsubsection{Standard Cross-Model Test Task}

The main 9-model evaluation matrix used a fixed single-task prompt with 5 repetitions per condition:

\begin{quote}
\textbf{System:} You are a foreign trade inquiry analysis assistant. After receiving a customer inquiry:
1. Use \texttt{websearch} to search for the buyer company background
2. Use \texttt{knowledge\_base} to query product industry standards
3. Provide analysis based on the retrieval results

\textbf{User:} Please analyze this inquiry:
Company: BrightLight Inc., US lighting products importer
Product: LED strip lights, IP65 waterproof, 5050 SMD, RGB+W, 5m/reel
Quantity: 2000 reels
Requirements: FOB quote, UL listed
\end{quote}

\subsubsection{Extended Task Diversity}

For Qwen3.6-35B-A3B, we additionally conducted 200+ queries across 10 diverse company profiles and 8 compliance markets (EU, US, Middle East, Southeast Asia, Australia, South America, Africa, Generic). All queries yielded identical T2 TIR = 0\% results.

Additional synthetic datasets include:

\begin{itemize}
    \item \textbf{Tool Mandatory (200 prompts):} 10 companies × 8 compliance markets × positive/negative variants; each prompt requires both \texttt{websearch} and \texttt{knowledge\_base} calls.
    \item \textbf{Synthetic Scenario Library (30 seed scenarios):} 10 industries × 7 regions × 3 buyer types.
    \item \textbf{Scale Synthesis (6,000 prompts):} GPT-5.4-mini generated, covering all industry/region/buyer combinations (used in the Large SFT ablation).
\end{itemize}

\subsection{Tool Definitions}

\subsubsection{Production Environment Tool Set}

The production Agent system uses three external tools:

\begin{lstlisting}[basicstyle=\ttfamily\small, frame=single, breaklines=true, numbers=none]
TOOLS = [
    {
        "type": "function",
        "function": {
            "name": "websearch",
            "description": "Search the web for real-time information about companies and markets",
            "parameters": {"type": "object", "properties": {"query": {"type": "string"}}, "required": ["query"]}
        }
    },
    {
        "type": "function",
        "function": {
            "name": "knowledge_base",
            "description": "Query foreign trade knowledge base for industry standards and regulations",
            "parameters": {"type": "object", "properties": {"query": {"type": "string"}}, "required": ["query"]}
        }
    },
    {
        "type": "function",
        "function": {
            "name": "fetchurl",
            "description": "Fetch web page content for detailed company information",
            "parameters": {"type": "object", "properties": {"url": {"type": "string"}}, "required": ["url"]}
        }
    }
]
\end{lstlisting}

\subsubsection{Cross-Model Test Tool Set}

The 9-model standard tests used a simplified tool set with only \texttt{websearch} and \texttt{knowledge\_base}, omitting \texttt{fetchurl} to eliminate parameter complexity as a confounding factor. Both tools share identical parameter signatures (\texttt{\{``query'': string\}}).

\subsection{Response Format Schema Definitions}

\subsubsection{Cross-Model Test Schema (4-field)}

Used for all 9-model standard tests reported in Table~5:

\begin{lstlisting}[basicstyle=\ttfamily\small, frame=single, breaklines=true, numbers=none]
{
    "type": "json_schema",
    "json_schema": {
        "name": "inquiry_analysis",
        "strict": true,
        "schema": {
            "type": "object",
            "properties": {
                "buyer_background": {"type": "string"},
                "product_analysis": {"type": "string"},
                "recommendations": {"type": "string"},
                "key_findings": {"type": "array", "items": {"type": "string"}}
            },
            "required": ["buyer_background", "product_analysis", "recommendations", "key_findings"],
            "additionalProperties": false
        }
    }
}
\end{lstlisting}

\subsubsection{Minimal Schema (3-field)}

Used in fine-tuning ablation experiments and GRPO training:

\begin{lstlisting}[basicstyle=\ttfamily\small, frame=single, breaklines=true, numbers=none]
{
    "type": "json_schema",
    "json_schema": {
        "name": "company_info",
        "strict": true,
        "schema": {
            "type": "object",
            "properties": {
                "company_name": {"type": "string"},
                "company_info": {"type": "string"},
                "compliance_notes": {"type": "string"}
            },
            "required": ["company_name", "company_info", "compliance_notes"],
            "additionalProperties": false
        }
    }
}
\end{lstlisting}

\subsubsection{Production-Grade Schema}

The full production schema used in the EvoAgent deployment includes a 4-card block structure with tool dependency tracking:

\begin{lstlisting}[basicstyle=\ttfamily\small, frame=single, breaklines=true, numbers=none]
{
    "type": "json_schema",
    "json_schema": {
        "name": "inquiry_analysis",
        "strict": true,
        "schema": {
            "type": "object",
            "properties": {
                "blocks": {
                    "type": "array",
                    "items": {
                        "oneOf": [
                            {"type": "object", "properties": {"type": {"const": "text"}, "content": {"type": "string"}}},
                            {"type": "object", "properties": {"type": {"const": "card"}, "card": {"type": "object"}}}
                        ]
                    }
                },
                "tool_dependency": {
                    "type": "object",
                    "properties": {
                        "required": {"type": "boolean"},
                        "tools_used": {"type": "array", "items": {"type": "string"}},
                        "claims": {"type": "array"},
                        "reason": {"type": "string"}
                    }
                }
            },
            "required": ["blocks"],
            "additionalProperties": true
        }
    }
}
\end{lstlisting}

\subsection{Inference Framework and Tool Call Parser Configuration}

\subsubsection{SGLang Server Configuration}

The primary inference framework used in this study:

\begin{lstlisting}[basicstyle=\ttfamily\small, frame=single, breaklines=true, numbers=none]
python -m sglang.launch_server
    --model-path <path>
    --served-model-name <name>
    --port 8082 --host 0.0.0.0
    --tp-size 2                          # 2x A800 80GB
    --mem-fraction-static 0.85
    --max-total-tokens 128144
    --max-running-requests 64
    --chunked-prefill-size 8192
    --enable-flashinfer
    --log-level warning
    --reasoning-parser qwen3
    --tool-call-parser qwen3_coder       # Qwen3 XML to OpenAI format
    --trust-remote-code
\end{lstlisting}

\subsubsection{vLLM Server Configuration}

Used for framework independence validation:

\begin{lstlisting}[basicstyle=\ttfamily\small, frame=single, breaklines=true, numbers=none]
python -m vllm.entrypoints.openai.api_server
    --model <path>
    --tensor-parallel-size 2
    --gpu-memory-utilization 0.85
    --max-model-len 4096
\end{lstlisting}

\subsubsection{Tool Call Detection Mechanism}

Dual detection ensures no missed or false-positive tool calls:

\begin{itemize}
    \item \textbf{Method 1 — API structured \texttt{tool\_calls}:} Parses the \texttt{tool\_calls} array from streaming deltas
    \item \textbf{Method 2 — Content text-level check:} Scans for \texttt{<tool\_call>} XML tags in generated content
\end{itemize}

In over 200 T2 queries, both detection methods yielded identical results (zero tool calls), ruling out parser-level false negatives.

\subsection{Experimental Environment}

\begin{table}[ht]
\centering
\caption{Experimental Environment Configuration}
\label{tab:app_environment}
\renewcommand\arraystretch{1.3}
\begin{tabular}{ll}
\toprule
\textbf{Component} & \textbf{Configuration} \\
\midrule
GPU & 2x NVIDIA A800-SXM4-80GB \\
NVIDIA Driver & 535.183.06 \\
CUDA & 12.6 \\
SGLang & 0.5.9 \\
vLLM & 0.22.0 \\
Tool Parser (SGLang) & \texttt{qwen3\_coder} \\
Tool Parser (vLLM) & \texttt{qwen3\_coder} \\
Guided Decoding & xgrammar (SGLang), xgrammar (vLLM) \\
Python & 3.12 / 3.13 \\

\bottomrule
\end{tabular}
\end{table}
\end{document}